\documentclass{article}

\usepackage{microtype}
\usepackage{graphicx}
\usepackage{subfigure}
\usepackage{booktabs} 
\graphicspath{{./}{./Figures/}}

\usepackage[utf8]{inputenc} 
\usepackage[T1]{fontenc}    
\usepackage[english]{babel}
\usepackage{multirow}
\usepackage{adjustbox}

\usepackage{hyperref}

\usepackage[accepted]{icml2023}

\usepackage{amsmath}
\usepackage{amssymb}
\usepackage{mathtools}
\usepackage{amsthm}
\usepackage{bm}

\usepackage[capitalize,noabbrev]{cleveref}

\theoremstyle{plain}

\theoremstyle{definition}

\theoremstyle{remark}

\usepackage[disable,textsize=tiny]{todonotes}

\icmltitlerunning{Training Deep Surrogate Models with Large Scale Online Learning}

\begin{document}

\twocolumn[
\icmltitle{Training Deep Surrogate Models with Large Scale Online Learning}


\icmlsetsymbol{equal}{*}

\begin{icmlauthorlist}
\icmlauthor{Lucas Meyer}{datamove,edf}
\icmlauthor{Marc Schouler}{datamove}
\icmlauthor{Robert A. Caulk}{datamove}
\icmlauthor{Alejandro Ribes}{edf}
\icmlauthor{Bruno Raffin}{datamove}
\end{icmlauthorlist}

\icmlaffiliation{datamove}{Univ. Grenoble Alpes, Inria, CNRS, Grenoble INP, LIG}
\icmlaffiliation{edf}{Industrial AI Laboratory SINCLAIR, EDF Lab Paris-Saclay}

\icmlcorrespondingauthor{Lucas Meyer}{lucas.meyer@inria.fr}

\icmlkeywords{Training Framework, Online Deep Learning, High Performance
Computing, Partial Differential Equations, Numerical Simulations, Surrogate
Models, Synthetic Data, Large Scale Dataset}

\vskip 0.3in
]



\printAffiliationsAndNotice{}  

\begin{abstract}

The spatiotemporal resolution of Partial Differential Equations (PDEs) plays important roles in the mathematical description of the world's physical phenomena. In general, scientists and engineers solve PDEs numerically by the use of computationally demanding solvers. Recently, deep learning algorithms have emerged as a viable alternative for obtaining fast solutions for PDEs. Models are usually trained on synthetic data generated by solvers, stored on disk and read back for training. This paper advocates that relying on a traditional static dataset to train these models does not allow the full benefit of the solver to be used as a data generator. It proposes an open source online training framework\footnote{Code and documentation are respectively available at \url{https://gitlab.inria.fr/melissa/melissa} and \url{https://melissa.gitlabpages.inria.fr/melissa/}} for deep surrogate models. The framework implements several levels of parallelism focused on simultaneously generating numerical simulations and training deep neural networks. This approach suppresses the I/O and storage bottleneck associated with disk-loaded datasets, and opens the way to training on significantly larger datasets. Experiments compare the offline and online training of four surrogate models, including state-of-the-art architectures.  Results indicate that exposing deep surrogate models to more dataset diversity, up to hundreds of GB, can increase model generalization capabilities. Fully connected neural networks, Fourier Neural Operator (FNO), and Message Passing PDE Solver prediction accuracy is improved by 68\%, 16\% and 7\%, respectively.

\end{abstract}

\section{Introduction}
\label{sec:introduction}
PDEs are powerful mathematical tools commonly
used to describe the dynamics of complex phenomena. Although these tools remain
foundational to scientific advancements in fluid dynamics, mechanics,
electromagnetism, biology, chemistry or geosciences, running classical PDEs
\emph{solvers}, based on numerical techniques such as Finite Elements, Finite
Volumes, or Spectral Methods, can be computationally prohibitive
\cite{burden2015numerical}.

Training deep surrogate models of PDEs emerges as an alternative and
complementary approach~\cite{stevens2020ai}. In most cases, scientists introduce
physical knowledge to machine learning algorithms by developing surrogate models
\cite{raissi2019physics} or shaping neural network architectures to mimic
traditional solvers
\cite{li2020fourier,pfaff2020learning,brandstetter2021message}. Other studies
developed models to improve the approximation obtained by a solver at a coarse
resolution, faster than it would take to run the solver directly at the desired
resolution \cite{um2020solver,kochkov2021machine}. Most deep-learning algorithms
proposed in the literature are supervised; they are trained on simulations
generated by the same solvers they intend to replace or accelerate. Meanwhile, training
deep surrogate models that can approximate the solution for a family of
parameterized PDEs, requires generating large datasets for a good coverage of the
parameter space. 

In its general form, a PDE describes the evolution of a quantity $u$ defined in
space and time as below: 

\begin{align*}
    u \colon \Omega \times T & \longrightarrow \Omega' \\
    (x, t) & \longmapsto u(x, t)
\end{align*}
\begin{align}
    \mathcal{D}u & = f \label{eq:domain}\\
    u(x, t = 0) & = u_0 \label{eq:IC}\\
    u(x \in \partial \Omega, t) & = u_{\partial \Omega} \label{eq:BC}
\end{align}

where $\Omega \subset \mathbb{R}^{d_\text{in}}$, $\Omega' \subset \mathbb{R}^{d_\text{out}}$ and $T \subset \mathbb{R}^{+}$.  $\mathcal{D}$ is a differential operator and  $f$ is referred to as a forcing term accounting for external forces applied to the system. $u_0$ and $u_{\partial \Omega}$ are respectively the initial and boundary conditions. All these variables contribute to the parametrization of PDEs and are represented by a vector $\bm{\lambda}$. It encompasses the variability in the boundary conditions, initial conditions, the geometry of the domain, or even the coefficients of the PDE itself appearing in the operator $\mathcal{D}$.

Most deep surrogate approaches currently found in the literature have been
applied to less complex PDEs focused on lower spatial dimensions or reduced
resolutions. To our knowledge, it is uncommon to address highly parameterized
problems. When done, it is difficult to introduce a wide range of parameters
that exhibit much diversity. As a result, these deep surrogates are not as
versatile as traditional solvers. The gap between the simplicity of the problems
considered by deep learning approaches and the complexity of common PDE-based
research and application can be partly explained by the difficulty in obtaining
comprehensive datasets \cite{brunton2020machine}. Indeed, the training of those
algorithms suffers the exact flaw that motivated their development: \emph{the
data generation with traditional solvers is a slow process with a high memory
and storage footprint}.

This paper proposes a framework to orchestrate large-scale parallelized data generation and training. The present approach suppresses the I/O and storage bottleneck associated with disk-loaded datasets in traditional training approaches. It enables a shift towards a potentially endless data generation only bounded by computational resource availability. Specifically, the presented framework: 

\begin{itemize}

\item{enables the complete automation of large-scale
workflow deployment encompassing the parallel execution of many solver instances
(which may also be parallelized) sending data directly to a parallel training
server;}

\item{mitigates the potential bias inherently caused by the direct streaming of
generated data to the training process;} 

\item{demonstrates increased generalization capacity for surrogate models, compared to a traditional "offline"
training based on the repetition of samples over several epochs.}

\end{itemize}

\section{Related Work}
\label{sec:literature}
\subsection{Deep Learning for Numerical Simulations}

The approaches applying deep learning to accelerate numerical simulations are
many and various \cite{brunton2020machine,hennigh2021nvidia,karniadakis2021physics}. Most of them involve supervised training, relying on the simulation output of traditional solvers. However, unsupervised models can be trained in a data-free regime \cite{raissi2019physics,sirignano2018dgm,wandel2020learning}.  They follow the elegant idea introduced by \citet{raissi2019physics}: the model directly predicts the quantity $u$ described by the PDE. This physics informed neural network (PINN) is subsequently trained by minimizing the residual of the equations at random collocation points: \autoref{eq:domain}, \autoref{eq:IC}, and \autoref{eq:BC} respectively on the domain, initial and boundary points. Though a promising approach, some studies have reported difficulties in training PINNs, even on simple problems \cite{krishnapriyan2021characterizing}. Some of these difficulties are alleviated by using relevant gradient computation techniques \cite{rackauckas2020universal}. PINNs training can also integrate generated data \cite{raissi2020hidden}. It has proven successful in reconstructing high-fidelity simulation by using a subset of data generated with a computationally intensive solver \cite{lucor2022simple}.

While most of the models are trained with the supervision of data obtained from traditional solvers, the data ingestion techniques distinguish direct prediction models from autoregressive ones. Direct prediction models take the time as an input to predict the corresponding state of $u$ (\autoref{eq:direct_time}).

\begin{equation}
    f_\theta(t) \approx u_t
    \label{eq:direct_time}
\end{equation}

\citet{raissi2019physics,lu2021learning} provide examples of such models. These direct approaches suffer from generalization when extrapolating to times outside of their training range.

Autoregressive models replicate the iterative process of traditional solvers:
they predict the next time step from previous ones
(\autoref{eq:autoregressive}).

\begin{equation}
    f_\theta(u_t) \approx u_{t+\delta t}
    \label{eq:autoregressive}
\end{equation}

Autoregressive surrogate models come in different flavors. If the spatial grid used by the solver is regular, each time step can be seen as an image, and classical deep learning algorithms like U-Net \cite{ronneberger2015u} can be employed \cite{wang2020towards}. However, most solvers rely on irregular grids. To overcome the dependency on a fixed grid, scientists proposed to work in the Fourier space \cite{li2020fourier}, while others rely on graph neural networks \cite{pfaff2020learning,brandstetter2021message}.

Autoregressive networks tend to accumulate errors along the trajectory they reconstruct. \citet{brandstetter2021message} introduced a \emph{push forward trick} to minimize these errors. The network is trained not on a single transition between two consecutive time steps, but rather on an extended rollout trajectory. This training procedure improves the stability of the predictions \cite{takamoto2022pdebench} but its implementation requires access to the full trajectories.

Prior works, especially involving super-resolution, already integrate the solver to the training loop \cite{um2020solver,rackauckas2020universal,kochkov2021machine}. Solvers are implemented using frameworks that support automatic differentiation. It allows the backpropagation through the solver during the training of the network. These \emph{inline} approaches remain sequential: reference simulations are generated beforehand. The \emph{online} characterization of the presented framework, which is also compatible with such approaches, denotes the generation of reference simulations occurring \emph{in parallel} to the training.

\subsection{Data Efficiency}

Deep learning has benefited from large datasets widely adopted by the community \cite{russakovsky2015imagenet,merity2017pointer}. Datasets dedicated to physics and numerical simulations are emerging \cite{otness2021extensible,takamoto2022pdebench,bonnet2022airfrans}. They are much welcomed as they allow to benchmark models on standard PDEs. Nonetheless, physical systems and PDEs exhibit patterns as diverse as the world itself. The sole Navier-Stokes equations present a rich variety of dynamics depending on the compressibility of the fluid, its monophasic or multiphasic nature, etc.  Dedicated solvers are even developed to tackle these specificities individually \cite{ferziger2002computational}. No dataset can encompass all the variety of physical phenomena. Consequently, training a model on a specific problem demands a dedicated
dataset.

In addition, as the PDE parameters vary it requires a massive data to capture all its richness. To improve the accuracy of solver simulations one can increase the grid resolution, which increases the size of individual trajectories. The size of a dataset composed of such trajectories may soon reach the current memory capacity of the hardware. Deep learning techniques to train a model with fewer data while maintaining the accuracy that would be obtained with a comprehensive dataset have been scarcely applied in the context of numerical simulations. \emph{Transfer learning} has been studied to solve the same PDEs at different resolutions \cite{chakraborty2021transfer}. \citet{satorras2021n} proposed to use \emph{equivariance} models to capture the symmetries of a problem that would otherwise require much more training data.

\subsection{Parallel Training Frameworks}

The growing size of datasets and models, along with the quest for always faster
training has driven the machine learning community towards a regular use of high
performance computing (HPC) \cite{sergeev2018horovod,You2020Large}. Today
support for multi-GPU parallel training, combining data and model parallelism,
has become common and integrated into major deep learning frameworks
\cite{li2020pytorch}.  Reciprocally, scientists working on numerical
simulations, the traditional users of HPC, have started considering deep
learning to accelerate their simulations. Workflows coupling large-scale and
parallel simulations with machine learning have started to emerge
\cite{peterson2022enabling}.  \citet{lee2021scalable,brace2022coupling} apply
such approach to molecular dynamics, \citet{stiller2022continual} to plasma
physics. Notice that having simulation runs in lockstep with training opens the
door to \emph{active learning} \cite{ren2021survey}. The parameterization of the
next set of simulations to run can be chosen according to the current state of
training.  These works explore early versions of adaptive training.

Deep reinforcement learning (Deep RL) relies on similar workflows where concurrent actors, running one simulation instance controlled by the current policy, are generating state/actions trajectories sent to the learner to compute a better policy. Frameworks running such workflows on clouds or supercomputers are available \cite{espeholt2018impala,horgan2018distributed,liang2018rllib}. But these frameworks are specialized for Deep RL, addressing its specificities like off-policy issues. Moreover, to our knowledge, Deep RL is working with simulation codes of intermediate complexity that run on a single node, potentially using several cores or one GPU \cite{berner2019dota}. The Deep RL frameworks do not support massively multi-node parallel solvers, typically parallelized with MPI and OpenMP, that we target here. Additionally, they do not support the direct streaming of data generated by external programs.

\section{Method}
\label{sec:method}
\begin{figure}[ht!]
    \centering
    \includegraphics[width=\columnwidth]{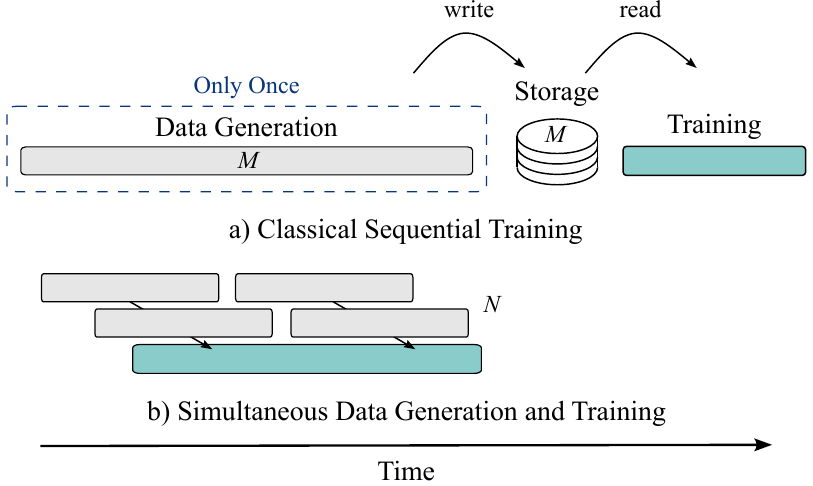}

    \caption{Classical I/O and storage intensive workflow for deep learning
    experiment (\textbf{a}) using $M$ samples presented several times through
    different epochs. (\textbf{b}) A file-free concurrent data generation and
    online training workflow processing $N$ samples. $N \gg M$.}

    \label{fig:sketch_concept}
\end{figure}

The traditional training from synthetic data consists in 1) generating the dataset 2) storing it on disk 3) reading it back for training (\autoref{fig:sketch_concept}). At scale, I/Os and storage become major constraints, limiting the size and representativeness of a training dataset.  We propose a file-avoiding framework that trains a deep surrogate model simultaneously on data streams generated from solver instances executed in parallel (\autoref{fig:sketch_concept}). The  I/Os and storage are drastically reduced. The training dataset becomes only limited by the amount of computing resources available. To put numbers in perspective, GPT-3 was trained on a main dataset of 570GB (plus 4 additional significantly smaller ones) \cite{brown2020language}, while the proposed workflow has been used for computing iteratively statistics for sensitivity analysis on 288TB of data generated and processed on-the-fly \cite{insitu-book-chapter:2022}. This opens the possibility to train on perpetually new data, while also allowing for sampling repetition as in classical training based on epochs.

\subsection{Framework Overview}

The framework is implemented as an extension of the open source project Melissa, initially designed to handle large-scale ensemble runs for sensitivity analysis \cite{terraz2017melissa} and data assimilation \cite{friedemann-melissaDA:2022}. The goal is to enable the online (no intermediate storage in files) training of a neural model from data generated through different simulation executions. These simulations are executed with different input parameters (parameter sweep), making the different members of the ensemble. We target executions on supercomputers where simulations can be large parallel solver codes executed on several nodes and the training parallelized on several GPUs. Compared to the original Melissa framework, the extension presented here provides deep learning support while addressing the specificities of online training, and implements new communication patterns.

The framework consists of three main elements: several clients, one server, and one launcher (\autoref{fig:melissa_framework}):

\begin{itemize}

  \item{\textbf{The clients} execute the same partial differential equations solver but
  each instance with a different set of parameters  $\bm{\lambda}$.  Once started, each
  client connects to the server and directly streams the generated data to the
  server.}

  \item{\textbf{The server} is in charge of the training. It receives data from connected
  clients. At any time, a subset of the entire data received so far is kept in a
  \emph{memory buffer}. Batches are assembled from this memory buffer to feed
  the network. The server also chooses the set of parameters $\bm{\lambda}$ for
  each client to run, and forwards them to the launcher.}
  
  \item{\textbf{The launcher} orchestrates and monitors the execution. On supercomputers,
  it requests resource allocations to the machine job scheduler (e.g. Slurm
  \cite{yoo2003slurm} or OAR \cite{capit2005batch}). On these allocations, the
  server starts first then the clients.} 

\end{itemize}

\definecolor{mypurple}{RGB}{101, 97, 169}
\definecolor{edfblue}{RGB}{9, 53, 112}
\begin{figure}[ht]

    \includegraphics[width=\columnwidth]{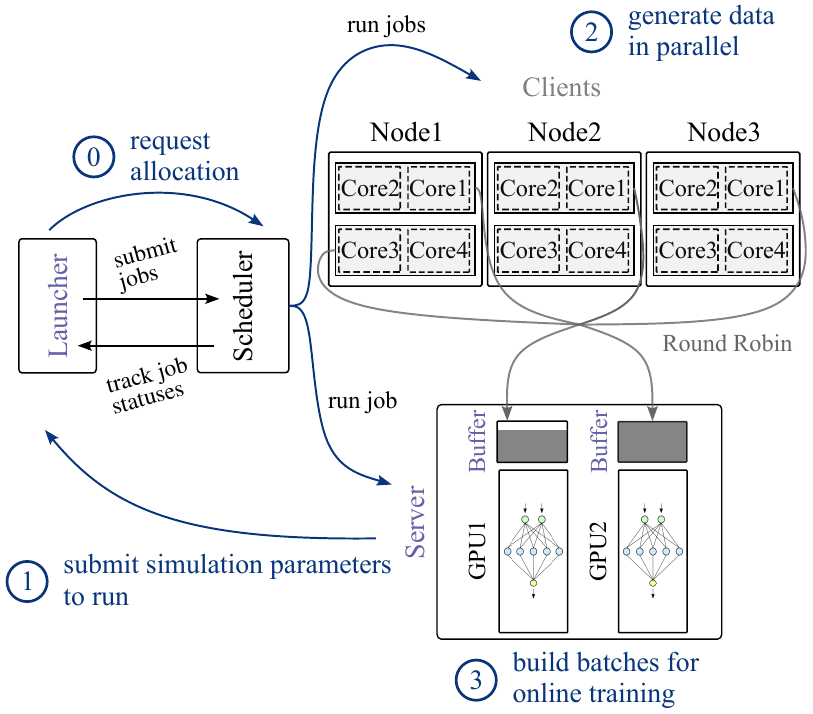}

    \caption{Framework architecture and workflow. The specific components introduced by the framework are colored in \textcolor{mypurple}{purple} (e.g. the \textcolor{mypurple}{server}). The different stages of the workflow are highlighted in \textcolor{edfblue}{blue}.  \textcolor{edfblue}{\textbf{2}} and \textcolor{edfblue}{\textbf{3}} occur simultaneously. In the depicted example, 6 clients run concurrently on 3 nodes.  They are themselves parallelized on 2 cores. The training is performed in a data-distributed manner on 2 GPUs.}
    \label{fig:melissa_framework}
  \end{figure}

\autoref{sec:appendix:framework} provides details about the framework implementation.

\subsection{Data Flow}
\label{sec:data_flow}

The server can be executed on several GPUs, located or not on the same nodes, for data parallel training. Each GPU is associated with its own memory buffer of fixed size.

When a new client starts a simulation, it establishes connections with each of the server processes. As soon as the running simulation produces new data, they are sent to the server. Data are distributed between GPUs in a Round Robin fashion. No data are ever stored on disk, hence avoiding costly I/O operations.

The paper focuses on one-way data transfer. The clients transmit all the data required for training: the field of interest $u_t$ and possibly additional information the solver can provide like the adjoint. Nonetheless, the framework can also support bidirectional communications where the server communicates data back to the clients \cite{friedemann-melissaDA:2022}. This is for instance necessary for applications based on backpropagation through the solver \cite{um2020solver,rackauckas2020universal} or Deep RL for which updates of the policy network must be passed to the actors.

\subsection{Data Management}
\label{sec:datamanagement}
\begin{figure}[h]
    \includegraphics[width=\columnwidth]{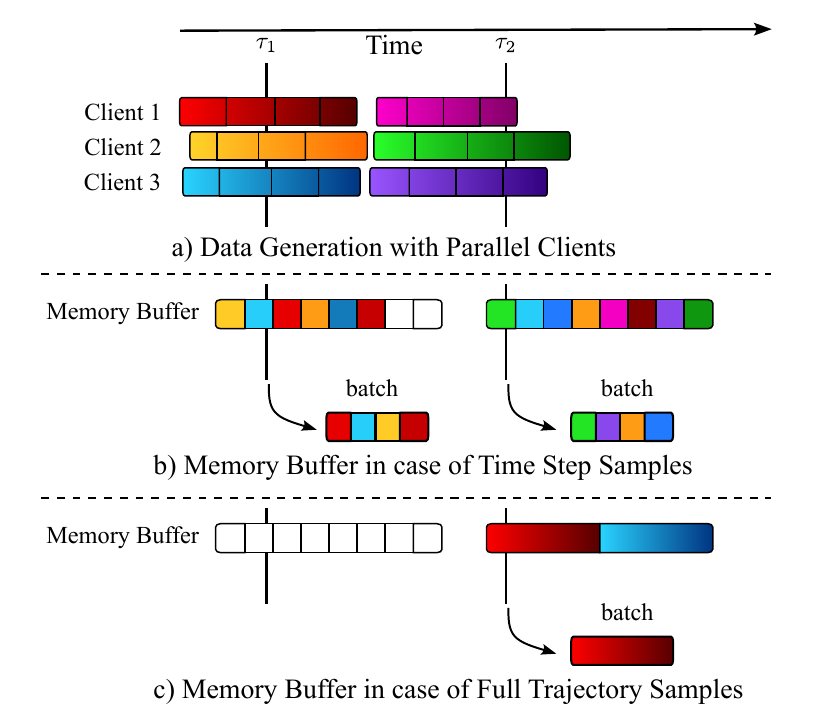}

    \caption{Flow of the generated simulation data. In \textbf{a}, 3 processes
    run 6 simulations in parallel. Groups of concurrent simulations are executed
    sequentially. In \textbf{b}, as soon as a simulation time step
    $u^t_{\lambda_i}$ is computed it is passed to the memory buffer. In
    \textbf{c}, the training requires access to the full trajectory and so the
    unit element in the buffer is a full trajectory.  The content of the memory
    buffer is repesented a two different times $\tau_1$ and  $\tau_2$.
    At time $\tau_1$ no batch can be extracted for  \textbf{c} as no full trajectory is  available yet.}
  \label{fig:memory_buffer}
\end{figure}

Assembling batches with enough diversity is critical to apply stochastic
gradient descent algorithm \cite{bottou2018optimization}.  Biased batches can
otherwise lead to catastrophic forgetting \cite{kemker2018measuring} marked by
decreased performances of the trained network.

The online training workflow comes with specific possible sources of bias: 
 \begin{itemize}
    \item{\textbf{Intra-simulation bias}: Solvers produce dynamics as time series by
    discretizing time and progressing iteratively. After $T$ time steps, only the data
    $u^t_{\lambda}, 0 \leq t \leq T $ for the simulation parameter $\bm{\lambda}$ are available
    for training.}

\item{\textbf{Inter-simulation bias}: Computational resources are limited and so often not
    all simulation instances can be executed concurrently. Let's assume that only
    $c$ simulations can be executed concurrently at any time.  At the end of the
    execution of the $c$ first simulations, training only has access to the data
    $u^t_{\lambda_i}, 0 < i \leq c$. }

    \item{\textbf{Memory bias}: The set of data available for training is not only limited to the already generated data, it is further reduced due to memory constraints. At most, a rolling subset of the generated data that fits the given memory budget can be kept and used for training.}

\end{itemize}

The management of the memory buffer is key to counteract these biases (\autoref{fig:memory_buffer}).  Different management policies are available ranging from simple queues to reservoir sampling \cite{efraimidis2006weighted}. The one used in the experiments of this paper allows writing incoming data on free slots only, while each (random) read suppresses the corresponding data. So each data is seen only once during training. A watermark is also used to prevent the data loader from extracting samples from the buffer when too few are present. This also ensures bigger diversity in batches randomly taken from the buffer population. 

The modality of training also impacts data management. The training may work with individual time steps, an interval of successive time steps or the full trajectory (\autoref{fig:memory_buffer}). Autoregressive models (\autoref{eq:autoregressive}) require at least two successive time steps: the input and the target of the model. However, the implementation of the \emph{push forward} trick introduced by \citet{brandstetter2021message} to stabilize long-term prediction of the network requires access to the full trajectory. This paper experiments with different situations. When the full trajectory is needed, the unit sample in the memory buffer encompasses the full trajectory.

The sampling strategy implemented by the server for selecting the parameters $\bm{\lambda}$ to run also has an impact on the inter-simulation bias. Classical Monte Carlo sampling is used throughout the paper. Nonetheless, the framework provides the abstraction to implement more advanced sampling strategies.

\section{Experiments}
\label{sec:experiments}

The experiments consist in evaluating the performance of the framework in training state-of-the-art deep surrogate architectures on classical PDEs.  \autoref{tab:experiments} lists the different experiments, with the model that is trained, the equations ruling the use case, and the source of the model or solver. Except stated otherwise, the hyper-parameters of the training (e.g. the batch size) are kept identical to the original paper. Data generation details are provided in \autoref{sec:appendix:equations}. Experiments are run on NVIDIA V100 GPUs and Intel Xeon 2.5GHz processors. \autoref{tab:training_comparison} synthesizes the experimental resources and the results obtained. 


\begin{table*}[ht]

    \caption{Combinations of models and PDE use cases in the different
    experiments performed with the online training framework.}

    \label{tab:experiments}
\vskip 0.15in
\begin{center}
\begin{small}
\begin{sc}
\begin{tabular}{cccc}
\toprule
Experiment & PDE & Model & Source \\
\midrule
E1 & Heat equation & Fully Connected & \textemdash \\
E2 & Lorenz's system & Fully Connected & \citet{lorenz1963deterministic} \\
E3 & Navier-Stokes & U-Net & \citet{takamoto2022pdebench} \\
E4 & Navier-Stokes & FNO & \citet{li2020fourier} \\
E5 & Mixed Advection-Diffusion & Message Passing PDE Solver & \citet{brandstetter2021message} \\
\bottomrule
\end{tabular}
\end{sc}
\end{small}
\end{center}
\vskip -0.1in
\end{table*}

\begin{table*}[ht!]
\caption{Comparison of the training modes for the different experiments. \textit{RESOURCES} column indicates computing resources used for data generation.}
\label{tab:training_comparison}
\vskip 0.15in
\begin{center}
\begin{small}
\begin{sc}
\begin{tabular}{lccccccc}
\toprule
\adjustbox{stack=cc}{Experiment \\ ~} & \adjustbox{stack=cc}{Resources \\ ~} & \adjustbox{stack=cc}{Generation \\ (hours)} & \adjustbox{stack=cc}{Total \\ (hours)} & \adjustbox{stack=cc}{Dataset\\ (GB)} & \adjustbox{stack=cc}{Unique Samples \\ ($N$)} & \adjustbox{stack=cc}{RMSE $\downarrow$ \\ ~} & \adjustbox{stack=cc}{Gain\\(\%)} \\
\midrule
    E1 offline & 10 cores & 0.079 & 0.284 & 0.08 & 1e4 & 2.46 & \textemdash\\
    E1 online & 10 cores & \textemdash & 0.710 & 80.0 & 1e6  & 0.766 & 68.9 \\
\midrule
    E3 offline & 10 cores & 3 & 3.82 & 0.328 & 1e3 & 0.1455 & \textemdash\\
    E3 online & 1200 cores & \textemdash & 6.39 & 328 & 5e5 & 0.1463 & -0.549\\
    E4 offline & 10 cores & 2.4 & 4.13 & 0.328 & 1e3 & 0.0876 & \textemdash\\ 
    E4 online & 160 GPUs & \textemdash & 3.95 & 328 & 5e5 & 0.0739 & 15.6 \\
\midrule
    E5  offline & 1 GPU & 7.41 & 9.80 & 1.17 & 2.048e3 & 7.64 & \textemdash\\
    E5 online & 400 Cores & \textemdash & 2.50 & 82.4 &  4.96e5 & 7.18 & 6.02 \\
\bottomrule
\end{tabular}
\end{sc}
\end{small}
\end{center}
\vskip -0.1in
\end{table*}

\subsection{Dataset Size Impact}
\label{sec:exp:bigger_datasets}

A first motivational experiment examines the impact of the number of
trajectories in the dataset on model performance  in a classical offline
training.

The \textbf{Message Passing Neural PDE Solver} is trained on the \textbf{mixed advection-diffusion} dataset consisting of a training set and a test set of respectively 2,048 and 128 trajectories as in \citet{brandstetter2021message}. The trajectories differ only by the initial conditions of the PDE ($\bm{\lambda} = u_0$). Smaller training datasets are extracted from the original one. For each of these datasets, the model is always trained on 640,000 batches, and tested on the same test set.  \autoref{tab:mp-pde-offline-dataset-size} shows the error decreases with the number of trajectories available.

\begin{table}[ht]

    \caption{Test error of the \textbf{Message Passing Neural PDE Solver} trained on datasets of the \textbf{mixed advection-diffusion} with different numbers of training trajectories.}

    \label{tab:mp-pde-offline-dataset-size}
\vskip 0.15in
\begin{center}
\begin{small}
\begin{sc}
\begin{tabular}{ccc}
\toprule
\# Training Simulations & Size (MB) & RMSE $\downarrow$ \\
\midrule

409 & 240 & 4.13\\
1024 & 600 & 2.55\\
1638 & 960 & 2.13\\
2048 & 117 & 1.89\\

\bottomrule
\end{tabular}
\end{sc}
\end{small}
\end{center}
\vskip -0.1in
\end{table}

\begin{figure}[ht]
    \includegraphics[width=\columnwidth]{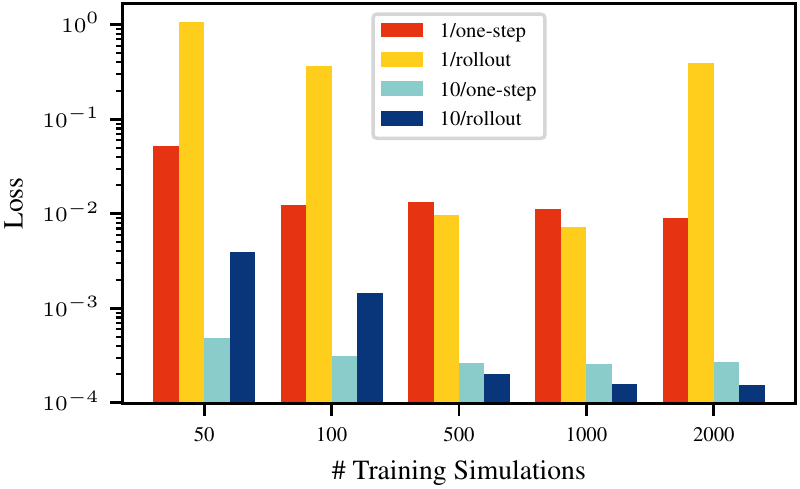}

    \caption{Test error of \textbf{FNO} trained on datasets of the
    \textbf{simple advection} with different numbers of trajectories. The
    different training strategies are noted \emph{history}/\emph{rollout
    length}.}

    \label{fig:fno-offline-dataset-size}
\end{figure}

\textbf{FNO} is trained with 5 datasets of different sizes from the
\textbf{simple advection} data. Trajectories differ by the initial conditions of
the equation ($\bm{\lambda} = u_0$). We also test different training strategies
combining an input of 1 or 10 consecutive time steps and a single time step or full
rollout prediction. \autoref{fig:fno-offline-dataset-size} shows the error of
the different trained models on the same test dataset of 100 unseen
trajectories.  All modalities benefit from larger datasets, except the history
of 1  and full rollout that likely suffers from overfitting at 2000
trajectories. The benefit of a larger dataset is particularly noticeable when
using the default configuration from \cite{li2020fourier} (history of 10 and a
full trajectory rollout) with a gain of $96\%$.

The experiment shows that state-of-the-art models make lower errors in inference when trained on more trajectories. It corroborates similar results obtained for \textbf{FNO} on the \textbf{Navier-Stokes equation} \cite{li2020fourier}. Although the datasets considered presently are still relatively small (up to 1.6 GB), generating much more trajectories for training, which would help generalization, would soon be prohibitive in terms of memory. The framework proposed in the present article addresses this difficulty by avoiding writing any data on the disk. 

\subsection{Online Learning with Single Step Samples}
\label{sec:exp:sing_step}
 
The heat equation example (\hyperref[tab:experiments]{E1}) trains a fully
connected network to predict the entire 2D temperature field, given six
temperature inputs: four boundary conditions, one initial condition, and the
time step ($\bm{\lambda} = [u_0, u_\text{BC}]$, each  randomly sampled between
100K and 500K).  The solver used to generate the data approximates the solution
on a 100 x 100 spatial grid for 100 time steps representing 1 second. It relies
on an implicit finite difference scheme. For the offline mode, two training
datasets of respectively 100 and 500 trajectories are generated.  The initial
and boundary temperature conditions vary for each trajectory with values
randomly selected between 100K and 500K. For the online mode, 1 node executes 5
clients to generate 10,000 trajectories. Each client runs the solver in parallel
on 2 cores. Each time a new time step is completed, the 2D temperature field is
sent to the memory buffer.

The model architecture consists of 3 layers of 1024 features, followed by ReLU activation function except for the output layer. The model performs direct prediction as described in \autoref{eq:direct_time}. Both online and offline training procedures follow the same learning rate schedule starting with a value of $1\text{E}^{-3}$ and decaying exponentially. For offline training, the number of epochs is adjusted to always train with 100,000 batches. The offline training dataset is repeated between epochs.

\begin{figure}[ht]
    \includegraphics[width=\columnwidth]{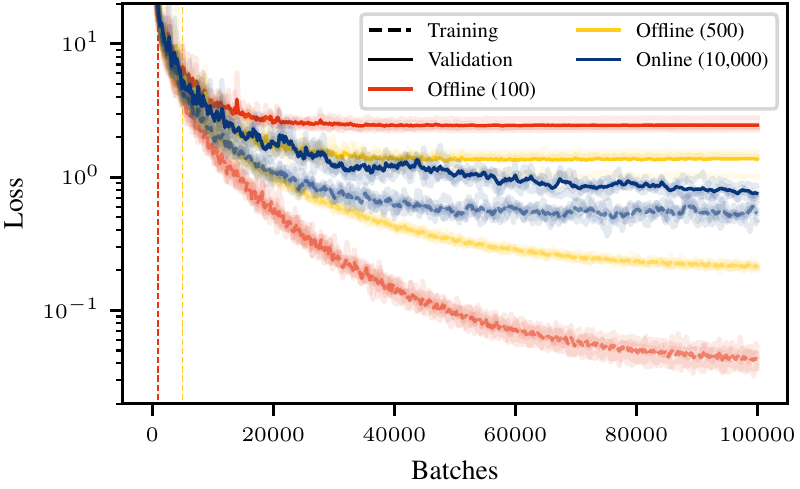}

    \caption{Comparison of the online and offline training on the \textbf{Heat Equation} (\hyperref[tab:experiments]{E1}). Solid lines represent the mean values of five realizations. Vertical lines represent the first epoch of the corresponding offline training, after which the dataset is repeated while the online setting always processes new data.}

    \label{fig:heatequation_loss}
\end{figure}

\autoref{fig:heatequation_loss} shows, for 5 repetitions of the experiment, the training and validation losses for the two offline training datasets and the online training.  The mean is plotted in solid lines and bold colors. The online training presents more diversity to the network which is correlated with better generalization. The online example enables a deeper exploration of the problem resulting in about $70\%$ improved validation error compared to 100 simulations and 100 epochs offline. Additionally, the gap between the decreasing training loss and the plateaued validation loss of the model trained in the offline settings is symptomatic of overfitting. The model trained with the online framework is less prone to such overfitting.

In the example of the \textbf{Lorenz's system} (\hyperref[tab:experiments]{E2}) the goal is to train a fully connected autoregressive model composed of 3 layers of 512 features followed by SiLU activation to recover the chaotic dynamics of the system. The model takes as input $\rho$, the parameter of the equation that varies across the simulations, and $X_t$ the position at the current time step ($\bm{\lambda} = [u_0, \rho]$). It predicts the next position $X_{t+ \delta t}$. The trajectory generation is detailed in \autoref{sec:appendix:lorenz}. Three different offline training datasets are used for the experiment. They respectively count 100, 10 and 10,000 trajectories. The trajectories of the latter are subsampled every 100 time steps. 10 and 10,000 datasets are symbolic of more complex systems for which storage limitations impose a scarcity of data.  For each dataset, the model is trained for 100, 1000, and 100 epochs respectively. For the online dataset, we generate 10000 trajectories. The $u_0$ parameter is randomly sampled.  In the \emph{streaming} online training, the memory buffer is a simple FIFO queue and the trajectories are generated from simulations with increasing values of $\rho$, creating an inter-simulation bias not altered by the FIFO queue.  The \emph{Sampling + Buffer} online training randomly samples $\rho$ and uses the default memory buffer introduced in \autoref{sec:datamanagement}. For all training strategies, the batch size is 1024. The validation dataset consists of 10 trajectories.

\begin{figure}[ht]
    \includegraphics[width=\columnwidth]{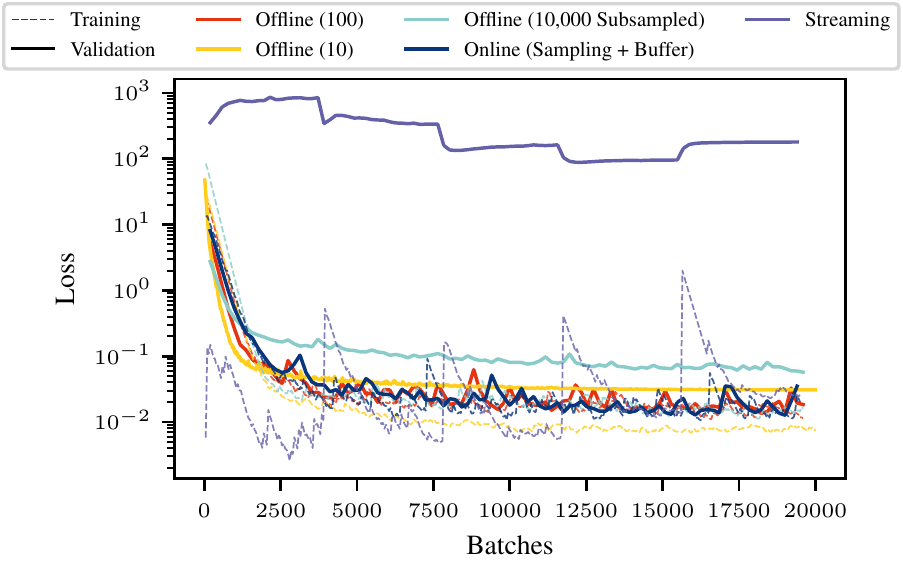}

    \caption{Comparison between the training and validation losses for different training strategies on the \textbf{Lorenz's system} (\hyperref[tab:experiments]{E2}).}

    \label{fig:lorenz_loss}

\end{figure}

\autoref{fig:lorenz_loss} shows the training dynamics for the different offline and online settings. The offline training with 10 trajectories and the one with 10,000 subsampled trajectories present signs of overfitting, indicating that more data are necessary to properly capture the chaotic dynamic of the system. Online training with 10,000 performs as well as offline training with 100 trajectories over 100 epochs. The \emph{streaming} training presents the worse validation curve which can be explained by catastrophic forgetting \cite{kemker2018measuring}. These results highlight the importance of the bias mitigation strategy and the ability of the framework to outperform classical training with scarce data. The efficiency of the bias mitigation strategy from the perspective of the batch statistics is further discussed in \autoref{sec:appendix:bias_mitigation}.

\subsection{Online Training on Full Trajectories}
\label{sec:exp:full_trajectories}

This subsection focuses on the online training of \textbf{FNO} (\hyperref[tab:experiments]{E4}), \textbf{U-Net} (\hyperref[tab:experiments]{E3}), and \textbf{Message Passing Neural PDE Solver} (\hyperref[tab:experiments]{E5}).
 
 The \textbf{FNO} (in its 2-D version) and \textbf{U-Net} architectures are
 trained on a use case ruled by the \textbf{Navier-Stokes equations}, as in \citet{li2020fourier,takamoto2022pdebench}.  For the offline mode, the training
 dataset consists of 1,000 trajectories repeated over 500 epochs with a batch size of 20 ($\bm{\lambda} = u_0$). For the online
 mode, data generation is either performed on 30 nodes of 40 cores each, or on
 40 nodes of 4 GPUs each (in this case clients run on GPU).  This allows running
 31,250 clients each generating 16 trajectories for a total of 500,000
 simulations. The same validation dataset is used to evaluate the performance of
 both training sessions. It consists of 200 trajectories generated offline. Both
 \textbf{FNO} and \textbf{U-Net} architectures are trained as autoregressive
 models following the original procedure \cite{li2020fourier}.

\begin{figure}[ht]
    \includegraphics[width=\columnwidth]{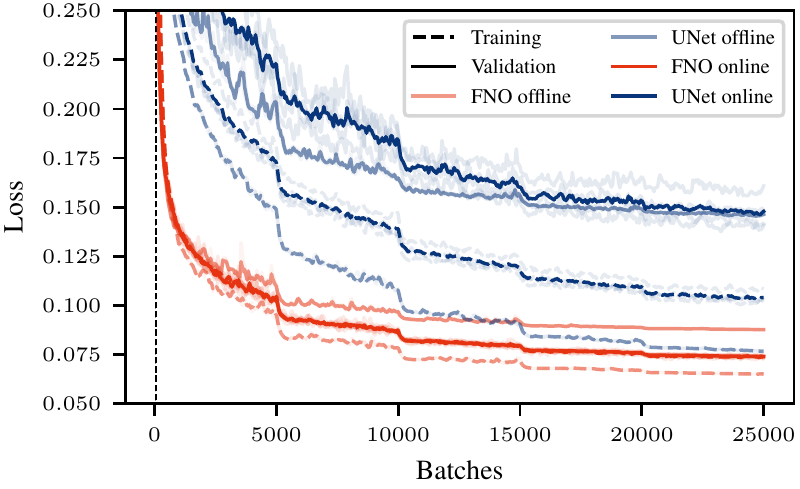}

    \caption{Comparison between the offline and online training of the \textbf{FNO} and \textbf{U-Net} architectures on the \textbf{Navier-Stokes Equations} (\hyperref[tab:experiments]{E3,E4}). The vertical line represents the first epoch accounting for 50 batches of 20 full trajectories each.}

    \label{fig:navierstokes_loss}
\end{figure}

\autoref{fig:navierstokes_loss} compares the training and validation loss for
the two architectures and for the two different ways of training. Each
experiment was repeated five times. For the \textbf{U-Net} architecture, the
online framework leads to similar results to offline training after 15,000
batches. On average, performances are similar but the online training is
characterized by strong variations around the mean value.  This instability
echoes the observations of \citet{takamoto2022pdebench}, where the so called
pushforward trick was used to stabilize the architecture. On the other hand, the
online framework has a clear positive impact on the performance of \textbf{FNO},
which outperforms U-Net. Online training is indeed characterized by a lower
validation loss compared to offline training.

The experiment \hyperref[tab:experiments]{E5} tests the impact of the online
training framework on the \textbf{Message Passing PDE Solver} with the use case
of the \textbf{mixed advection-diffusion} ($\bm{\lambda} = [u_0, \alpha, \beta,
\gamma]$). The solver generates trajectories by groups of 4. The equation
parameters $\alpha$, $\beta$, and $\gamma$ vary from one group to another. Only
the initial conditions vary within a group.  For offline data generation, the
solver is used as it is. For the online mode, it is only modified to send the
data to the server relying on the framework API, instead of writing the data on
disk. The offline training dataset consists of 2,048 trajectories. The model is
trained for 20 epochs. The online data generation is performed over 10 nodes of
40 cores each. It executes sequentially two ensembles of 400 clients. Each
client generates 128 batches of 4 simulations. In total, online data generation
produces 409,600 trajectories.  Online training is achieved on 2 GPUs (data
parallelism).  Clients feed iteratively each memory buffer associated with each
GPU in a Round Robin fashion. Both training methods are validated on the same
dataset consisting of 128 trajectories.

\begin{figure}[ht]

    \includegraphics[width=\columnwidth]{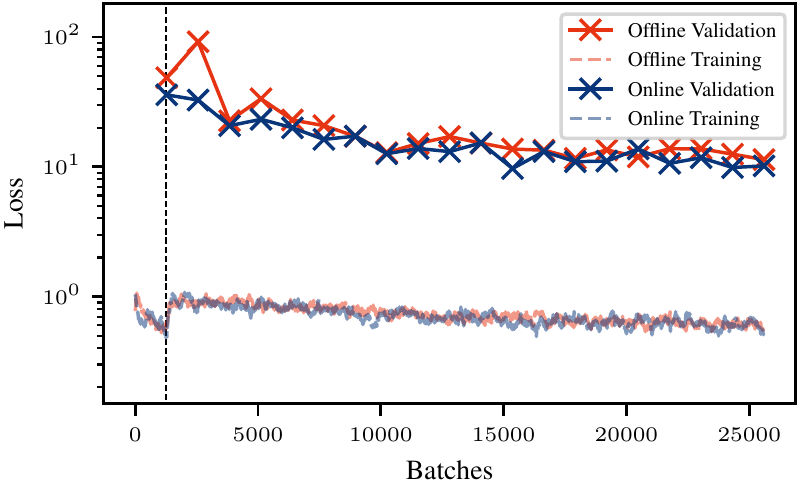}

    \caption{Comparison between the offline and online training of the \textbf{Message Passing PDE Server} on the \textbf{mixed advection-diffusion} (\hyperref[tab:experiments]{E5}). The vertical line represents the limit of the first epoch.}

    \label{fig:mp-pde_loss}
\end{figure}

\autoref{fig:mp-pde_loss} compares the training and validation losses. Online training presents an almost constantly lower validation loss. Evaluation on a test set shows that training the model with the online framework improves the performance by 7\% compared to the offline training.

\autoref{sec:exp:bigger_datasets} showed that, offline, bigger datasets generally improve surrogate model performances. The different online experiments indicate the framework sustains this trend despite the online setting. During training, it exposes the model to more unique samples which results in better generalization compared to offline training for the same number of processed batches. The different experiments also exhibit the versatility of the framework. Several models are trained on multiple PDE use cases. Several degrees of parallelism are presented for the data generation.





\section{Conclusion}
\label{sec:conclusion}
Deep surrogate models are promising candidates to accelerate the numerical simulations of PDEs, unlocking faster engineering processes and further scientific discoveries. The proposed online training framework leverages HPC resources to expose these models to bigger datasets, with more diverse trajectories, than it would otherwise be possible with classical offline training, due to I/O and storage limitations. By relying on simple techniques like a memory buffer and Monte Carlo sampling, the framework mitigates the inherent bias of streaming learning. Our framework enabled to process online datasets up to 328 GB and improve the generalization of fully connected neural networks by 68\%, FNO by 16\%, and Message Passing PDE Solver by 7\%. In future work, such improvement could be further enhanced by relying on active learning.


\section*{Acknowledgement}
This work was performed using HPC/AI resources from GENCI-IDRIS (Grant 2022-[AD010610366R1]),
and received funding from the European High-Performance Computing Joint Undertaking (JU) under grant agreement No 956560. 

\bibliography{references}

\begin{thebibliography}{51}
\providecommand{\natexlab}[1]{#1}
\providecommand{\url}[1]{\texttt{#1}}
\expandafter\ifx\csname urlstyle\endcsname\relax
  \providecommand{\doi}[1]{doi: #1}\else
  \providecommand{\doi}{doi: \begingroup \urlstyle{rm}\Url}\fi

\bibitem[Berner et~al.(2019)Berner, Brockman, Chan, Cheung, D{\k{e}}biak,
  Dennison, Farhi, Fischer, Hashme, Hesse, et~al.]{berner2019dota}
Berner, C., Brockman, G., Chan, B., Cheung, V., D{\k{e}}biak, P., Dennison, C.,
  Farhi, D., Fischer, Q., Hashme, S., Hesse, C., et~al.
\newblock Dota 2 with large scale deep reinforcement learning.
\newblock \emph{ArXiv preprint}, abs/1912.06680, 2019.

\bibitem[Bonnet et~al.(2022)Bonnet, Mazari, Cinnella, and
  Gallinari]{bonnet2022airfrans}
Bonnet, F., Mazari, J.~A., Cinnella, P., and Gallinari, P.
\newblock Airf{RANS}: High fidelity computational fluid dynamics dataset for
  approximating reynolds-averaged navier{\textendash}stokes solutions.
\newblock In \emph{Thirty-sixth Conference on Neural Information Processing
  Systems Datasets and Benchmarks Track}, 2022.

\bibitem[Bottou et~al.(2018)Bottou, Curtis, and
  Nocedal]{bottou2018optimization}
Bottou, L., Curtis, F.~E., and Nocedal, J.
\newblock Optimization methods for large-scale machine learning.
\newblock \emph{Siam Review}, 60\penalty0 (2):\penalty0 223--311, 2018.

\bibitem[Brace et~al.(2022)Brace, Yakushin, Ma, Trifan, Munson, Foster,
  Ramanathan, Lee, Turilli, and Jha]{brace2022coupling}
Brace, A., Yakushin, I., Ma, H., Trifan, A., Munson, T., Foster, I.,
  Ramanathan, A., Lee, H., Turilli, M., and Jha, S.
\newblock Coupling streaming ai and hpc ensembles to achieve 100--1000$\times$
  faster biomolecular simulations.
\newblock In \emph{2022 IEEE International Parallel and Distributed Processing
  Symposium (IPDPS)}, pp.\  806--816. IEEE, 2022.

\bibitem[Brandstetter et~al.(2022)Brandstetter, Worrall, and
  Welling]{brandstetter2021message}
Brandstetter, J., Worrall, D.~E., and Welling, M.
\newblock Message passing neural {PDE} solvers.
\newblock In \emph{The Tenth International Conference on Learning
  Representations, {ICLR} 2022, Virtual Event, April 25-29, 2022}, 2022.

\bibitem[Brown et~al.(2020)Brown, Mann, Ryder, Subbiah, Kaplan, Dhariwal,
  Neelakantan, Shyam, Sastry, Askell, Agarwal, Herbert{-}Voss, Krueger,
  Henighan, Child, Ramesh, Ziegler, Wu, Winter, Hesse, Chen, Sigler, Litwin,
  Gray, Chess, Clark, Berner, McCandlish, Radford, Sutskever, and
  Amodei]{brown2020language}
Brown, T.~B., Mann, B., Ryder, N., Subbiah, M., Kaplan, J., Dhariwal, P.,
  Neelakantan, A., Shyam, P., Sastry, G., Askell, A., Agarwal, S.,
  Herbert{-}Voss, A., Krueger, G., Henighan, T., Child, R., Ramesh, A.,
  Ziegler, D.~M., Wu, J., Winter, C., Hesse, C., Chen, M., Sigler, E., Litwin,
  M., Gray, S., Chess, B., Clark, J., Berner, C., McCandlish, S., Radford, A.,
  Sutskever, I., and Amodei, D.
\newblock Language models are few-shot learners.
\newblock In Larochelle, H., Ranzato, M., Hadsell, R., Balcan, M., and Lin, H.
  (eds.), \emph{Advances in Neural Information Processing Systems 33: Annual
  Conference on Neural Information Processing Systems 2020, NeurIPS 2020,
  December 6-12, 2020, virtual}, 2020.

\bibitem[Brunton et~al.(2020)Brunton, Noack, and
  Koumoutsakos]{brunton2020machine}
Brunton, S.~L., Noack, B.~R., and Koumoutsakos, P.
\newblock Machine learning for fluid mechanics.
\newblock \emph{Annual review of fluid mechanics}, 52:\penalty0 477--508, 2020.

\bibitem[Burden et~al.(2015)Burden, Faires, and Burden]{burden2015numerical}
Burden, R.~L., Faires, J.~D., and Burden, A.~M.
\newblock \emph{Numerical analysis}.
\newblock Cengage learning, 2015.

\bibitem[Capit et~al.(2005)Capit, Da~Costa, Georgiou, Huard, Martin,
  Mouni{\'e}, Neyron, and Richard]{capit2005batch}
Capit, N., Da~Costa, G., Georgiou, Y., Huard, G., Martin, C., Mouni{\'e}, G.,
  Neyron, P., and Richard, O.
\newblock A batch scheduler with high level components.
\newblock In \emph{CCGrid 2005. IEEE International Symposium on Cluster
  Computing and the Grid, 2005.}, volume~2, pp.\  776--783. IEEE, 2005.

\bibitem[Chakraborty(2021)]{chakraborty2021transfer}
Chakraborty, S.
\newblock Transfer learning based multi-fidelity physics informed deep neural
  network.
\newblock \emph{Journal of Computational Physics}, 426:\penalty0 109942, 2021.

\bibitem[Efraimidis \& Spirakis(2006)Efraimidis and
  Spirakis]{efraimidis2006weighted}
Efraimidis, P.~S. and Spirakis, P.~G.
\newblock Weighted random sampling with a reservoir.
\newblock \emph{Information processing letters}, 97\penalty0 (5):\penalty0
  181--185, 2006.

\bibitem[Espeholt et~al.(2018)Espeholt, Soyer, Munos, Simonyan, Mnih, Ward,
  Doron, Firoiu, Harley, Dunning, Legg, and Kavukcuoglu]{espeholt2018impala}
Espeholt, L., Soyer, H., Munos, R., Simonyan, K., Mnih, V., Ward, T., Doron,
  Y., Firoiu, V., Harley, T., Dunning, I., Legg, S., and Kavukcuoglu, K.
\newblock {IMPALA:} scalable distributed deep-rl with importance weighted
  actor-learner architectures.
\newblock In Dy, J.~G. and Krause, A. (eds.), \emph{Proceedings of the 35th
  International Conference on Machine Learning, {ICML} 2018,
  Stockholmsm{\"{a}}ssan, Stockholm, Sweden, July 10-15, 2018}, volume~80 of
  \emph{Proceedings of Machine Learning Research}, pp.\  1406--1415. {PMLR},
  2018.

\bibitem[Ferziger et~al.(2002)Ferziger, Peri{\'c}, and
  Street]{ferziger2002computational}
Ferziger, J.~H., Peri{\'c}, M., and Street, R.~L.
\newblock \emph{Computational methods for fluid dynamics}, volume~3.
\newblock Springer, 2002.

\bibitem[Friedemann \& Raffin(2022)Friedemann and
  Raffin]{friedemann-melissaDA:2022}
Friedemann, S. and Raffin, B.
\newblock {An elastic framework for ensemble-based large-scale data
  assimilation}.
\newblock \emph{{The international journal of high performance computing
  applications}}, 36\penalty0 (4):\penalty0 543--563, 2022.

\bibitem[Hennigh et~al.(2021)Hennigh, Narasimhan, Nabian, Subramaniam,
  Tangsali, Fang, Rietmann, Byeon, and Choudhry]{hennigh2021nvidia}
Hennigh, O., Narasimhan, S., Nabian, M.~A., Subramaniam, A., Tangsali, K.,
  Fang, Z., Rietmann, M., Byeon, W., and Choudhry, S.
\newblock Nvidia simnet™: An ai-accelerated multi-physics simulation
  framework.
\newblock In \emph{International Conference on Computational Science}, pp.\
  447--461. Springer, 2021.

\bibitem[Hintjens(2013)]{hintjens2013zeromq}
Hintjens, P.
\newblock \emph{ZeroMQ: messaging for many applications}.
\newblock " O'Reilly Media, Inc.", 2013.

\bibitem[Horgan et~al.(2018)Horgan, Quan, Budden, Barth{-}Maron, Hessel, van
  Hasselt, and Silver]{horgan2018distributed}
Horgan, D., Quan, J., Budden, D., Barth{-}Maron, G., Hessel, M., van Hasselt,
  H., and Silver, D.
\newblock Distributed prioritized experience replay.
\newblock In \emph{6th International Conference on Learning Representations,
  {ICLR} 2018, Vancouver, BC, Canada, April 30 - May 3, 2018, Conference Track
  Proceedings}, 2018.

\bibitem[Karniadakis et~al.(2021)Karniadakis, Kevrekidis, Lu, Perdikaris, Wang,
  and Yang]{karniadakis2021physics}
Karniadakis, G.~E., Kevrekidis, I.~G., Lu, L., Perdikaris, P., Wang, S., and
  Yang, L.
\newblock Physics-informed machine learning.
\newblock \emph{Nature Reviews Physics}, 3\penalty0 (6):\penalty0 422--440,
  2021.

\bibitem[Kemker et~al.(2018)Kemker, McClure, Abitino, Hayes, and
  Kanan]{kemker2018measuring}
Kemker, R., McClure, M., Abitino, A., Hayes, T.~L., and Kanan, C.
\newblock Measuring catastrophic forgetting in neural networks.
\newblock In McIlraith, S.~A. and Weinberger, K.~Q. (eds.), \emph{Proceedings
  of the Thirty-Second {AAAI} Conference on Artificial Intelligence, (AAAI-18),
  the 30th innovative Applications of Artificial Intelligence (IAAI-18), and
  the 8th {AAAI} Symposium on Educational Advances in Artificial Intelligence
  (EAAI-18), New Orleans, Louisiana, USA, February 2-7, 2018}, pp.\
  3390--3398. {AAAI} Press, 2018.

\bibitem[Kochkov et~al.(2021)Kochkov, Smith, Alieva, Wang, Brenner, and
  Hoyer]{kochkov2021machine}
Kochkov, D., Smith, J.~A., Alieva, A., Wang, Q., Brenner, M.~P., and Hoyer, S.
\newblock Machine learning--accelerated computational fluid dynamics.
\newblock \emph{Proceedings of the National Academy of Sciences}, 118\penalty0
  (21):\penalty0 e2101784118, 2021.

\bibitem[Krishnapriyan et~al.(2021)Krishnapriyan, Gholami, Zhe, Kirby, and
  Mahoney]{krishnapriyan2021characterizing}
Krishnapriyan, A.~S., Gholami, A., Zhe, S., Kirby, R.~M., and Mahoney, M.~W.
\newblock Characterizing possible failure modes in physics-informed neural
  networks.
\newblock In Ranzato, M., Beygelzimer, A., Dauphin, Y.~N., Liang, P., and
  Vaughan, J.~W. (eds.), \emph{Advances in Neural Information Processing
  Systems 34: Annual Conference on Neural Information Processing Systems 2021,
  NeurIPS 2021, December 6-14, 2021, virtual}, pp.\  26548--26560, 2021.

\bibitem[Lee et~al.(2021)Lee, Merzky, Tan, Titov, Turilli, Alfe, Bhati, Brace,
  Clyde, Coveney, et~al.]{lee2021scalable}
Lee, H., Merzky, A., Tan, L., Titov, M., Turilli, M., Alfe, D., Bhati, A.,
  Brace, A., Clyde, A., Coveney, P., et~al.
\newblock Scalable hpc \& ai infrastructure for covid-19 therapeutics.
\newblock In \emph{Proceedings of the Platform for Advanced Scientific
  Computing Conference}, pp.\  1--13, 2021.

\bibitem[Li et~al.(2020)Li, Zhao, Varma, Salpekar, Noordhuis, Li, Paszke,
  Smith, Vaughan, Damania, et~al.]{li2020pytorch}
Li, S., Zhao, Y., Varma, R., Salpekar, O., Noordhuis, P., Li, T., Paszke, A.,
  Smith, J., Vaughan, B., Damania, P., et~al.
\newblock Pytorch distributed: Experiences on accelerating data parallel
  training.
\newblock \emph{Proceedings of the VLDB Endowment}, 13\penalty0 (12), 2020.

\bibitem[Li et~al.(2021)Li, Kovachki, Azizzadenesheli, Liu, Bhattacharya,
  Stuart, and Anandkumar]{li2020fourier}
Li, Z., Kovachki, N.~B., Azizzadenesheli, K., Liu, B., Bhattacharya, K.,
  Stuart, A.~M., and Anandkumar, A.
\newblock Fourier neural operator for parametric partial differential
  equations.
\newblock In \emph{9th International Conference on Learning Representations,
  {ICLR} 2021, Virtual Event, Austria, May 3-7, 2021}, 2021.

\bibitem[Liang et~al.(2018)Liang, Liaw, Nishihara, Moritz, Fox, Goldberg,
  Gonzalez, Jordan, and Stoica]{liang2018rllib}
Liang, E., Liaw, R., Nishihara, R., Moritz, P., Fox, R., Goldberg, K.,
  Gonzalez, J., Jordan, M.~I., and Stoica, I.
\newblock Rllib: Abstractions for distributed reinforcement learning.
\newblock In Dy, J.~G. and Krause, A. (eds.), \emph{Proceedings of the 35th
  International Conference on Machine Learning, {ICML} 2018,
  Stockholmsm{\"{a}}ssan, Stockholm, Sweden, July 10-15, 2018}, volume~80 of
  \emph{Proceedings of Machine Learning Research}, pp.\  3059--3068. {PMLR},
  2018.

\bibitem[Lorenz(1963)]{lorenz1963deterministic}
Lorenz, E.~N.
\newblock Deterministic nonperiodic flow.
\newblock \emph{Journal of atmospheric sciences}, 20\penalty0 (2):\penalty0
  130--141, 1963.

\bibitem[Lu et~al.(2021)Lu, Jin, Pang, Zhang, and Karniadakis]{lu2021learning}
Lu, L., Jin, P., Pang, G., Zhang, Z., and Karniadakis, G.~E.
\newblock Learning nonlinear operators via deeponet based on the universal
  approximation theorem of operators.
\newblock \emph{Nature Machine Intelligence}, 3\penalty0 (3):\penalty0
  218--229, 2021.

\bibitem[Lucor et~al.(2022)Lucor, Agrawal, and Sergent]{lucor2022simple}
Lucor, D., Agrawal, A., and Sergent, A.
\newblock Simple computational strategies for more effective physics-informed
  neural networks modeling of turbulent natural convection.
\newblock \emph{Journal of Computational Physics}, 456:\penalty0 111022, 2022.

\bibitem[Merity et~al.(2017)Merity, Xiong, Bradbury, and
  Socher]{merity2017pointer}
Merity, S., Xiong, C., Bradbury, J., and Socher, R.
\newblock Pointer sentinel mixture models.
\newblock In \emph{5th International Conference on Learning Representations,
  {ICLR} 2017, Toulon, France, April 24-26, 2017, Conference Track
  Proceedings}, 2017.

\bibitem[Otness et~al.(2021)Otness, Gjoka, Bruna, Panozzo, Peherstorfer,
  Schneider, and Zorin]{otness2021extensible}
Otness, K., Gjoka, A., Bruna, J., Panozzo, D., Peherstorfer, B., Schneider, T.,
  and Zorin, D.
\newblock An extensible benchmark suite for learning to simulate physical
  systems.
\newblock In \emph{Thirty-fifth Conference on Neural Information Processing
  Systems Datasets and Benchmarks Track (Round 1)}, 2021.

\bibitem[Peterson et~al.(2022)Peterson, Bay, Koning, Robinson, Semler, White,
  Anirudh, Athey, Bremer, Di~Natale, et~al.]{peterson2022enabling}
Peterson, J.~L., Bay, B., Koning, J., Robinson, P., Semler, J., White, J.,
  Anirudh, R., Athey, K., Bremer, P.-T., Di~Natale, F., et~al.
\newblock Enabling machine learning-ready hpc ensembles with merlin.
\newblock \emph{Future Generation Computer Systems}, 131:\penalty0 255--268,
  2022.

\bibitem[Pfaff et~al.(2021)Pfaff, Fortunato, Sanchez{-}Gonzalez, and
  Battaglia]{pfaff2020learning}
Pfaff, T., Fortunato, M., Sanchez{-}Gonzalez, A., and Battaglia, P.~W.
\newblock Learning mesh-based simulation with graph networks.
\newblock In \emph{9th International Conference on Learning Representations,
  {ICLR} 2021, Virtual Event, Austria, May 3-7, 2021}, 2021.

\bibitem[Rackauckas et~al.(2020)Rackauckas, Ma, Martensen, Warner, Zubov,
  Supekar, Skinner, Ramadhan, and Edelman]{rackauckas2020universal}
Rackauckas, C., Ma, Y., Martensen, J., Warner, C., Zubov, K., Supekar, R.,
  Skinner, D., Ramadhan, A., and Edelman, A.
\newblock Universal differential equations for scientific machine learning.
\newblock \emph{ArXiv preprint}, abs/2001.04385, 2020.

\bibitem[Raissi et~al.(2019)Raissi, Perdikaris, and
  Karniadakis]{raissi2019physics}
Raissi, M., Perdikaris, P., and Karniadakis, G.~E.
\newblock Physics-informed neural networks: A deep learning framework for
  solving forward and inverse problems involving nonlinear partial differential
  equations.
\newblock \emph{Journal of Computational physics}, 378:\penalty0 686--707,
  2019.

\bibitem[Raissi et~al.(2020)Raissi, Yazdani, and Karniadakis]{raissi2020hidden}
Raissi, M., Yazdani, A., and Karniadakis, G.~E.
\newblock Hidden fluid mechanics: Learning velocity and pressure fields from
  flow visualizations.
\newblock \emph{Science}, 367\penalty0 (6481):\penalty0 1026--1030, 2020.

\bibitem[Ren et~al.(2021)Ren, Xiao, Chang, Huang, Li, Gupta, Chen, and
  Wang]{ren2021survey}
Ren, P., Xiao, Y., Chang, X., Huang, P.-Y., Li, Z., Gupta, B.~B., Chen, X., and
  Wang, X.
\newblock A survey of deep active learning.
\newblock \emph{ACM computing surveys (CSUR)}, 54\penalty0 (9):\penalty0 1--40,
  2021.

\bibitem[Rib{\'e}s et~al.(2022)Rib{\'e}s, Terraz, Fournier, Iooss, and
  Raffin]{insitu-book-chapter:2022}
Rib{\'e}s, A., Terraz, T., Fournier, Y., Iooss, B., and Raffin, B.
\newblock \emph{Unlocking Large Scale Uncertainty Quantification with In
  Transit Iterative Statistics}, pp.\  113--136.
\newblock Springer International Publishing, Cham, 2022.
\newblock ISBN 978-3-030-81627-8.

\bibitem[Ronneberger et~al.(2015)Ronneberger, Fischer, and
  Brox]{ronneberger2015u}
Ronneberger, O., Fischer, P., and Brox, T.
\newblock U-net: Convolutional networks for biomedical image segmentation.
\newblock In \emph{International Conference on Medical image computing and
  computer-assisted intervention}, pp.\  234--241. Springer, 2015.

\bibitem[Russakovsky et~al.(2015)Russakovsky, Deng, Su, Krause, Satheesh, Ma,
  Huang, Karpathy, Khosla, Bernstein, et~al.]{russakovsky2015imagenet}
Russakovsky, O., Deng, J., Su, H., Krause, J., Satheesh, S., Ma, S., Huang, Z.,
  Karpathy, A., Khosla, A., Bernstein, M., et~al.
\newblock Imagenet large scale visual recognition challenge.
\newblock \emph{International journal of computer vision}, 115\penalty0
  (3):\penalty0 211--252, 2015.

\bibitem[Satorras et~al.(2021)Satorras, Hoogeboom, and Welling]{satorras2021n}
Satorras, V.~G., Hoogeboom, E., and Welling, M.
\newblock E(n) equivariant graph neural networks.
\newblock In Meila, M. and Zhang, T. (eds.), \emph{Proceedings of the 38th
  International Conference on Machine Learning, {ICML} 2021, 18-24 July 2021,
  Virtual Event}, volume 139 of \emph{Proceedings of Machine Learning
  Research}, pp.\  9323--9332. {PMLR}, 2021.

\bibitem[Sergeev \& Del~Balso(2018)Sergeev and Del~Balso]{sergeev2018horovod}
Sergeev, A. and Del~Balso, M.
\newblock Horovod: fast and easy distributed deep learning in tensorflow.
\newblock \emph{ArXiv preprint}, abs/1802.05799, 2018.

\bibitem[Sirignano \& Spiliopoulos(2018)Sirignano and
  Spiliopoulos]{sirignano2018dgm}
Sirignano, J. and Spiliopoulos, K.
\newblock Dgm: A deep learning algorithm for solving partial differential
  equations.
\newblock \emph{Journal of computational physics}, 375:\penalty0 1339--1364,
  2018.

\bibitem[Stevens et~al.(2020)Stevens, Taylor, Nichols, Maccabe, Yelick, and
  Brown]{stevens2020ai}
Stevens, R., Taylor, V., Nichols, J., Maccabe, A.~B., Yelick, K., and Brown, D.
\newblock Ai for science: Report on the department of energy (doe) town halls
  on artificial intelligence (ai) for science.
\newblock Technical report, Argonne National Lab.(ANL), Argonne, IL (United
  States), 2020.

\bibitem[Stiller et~al.(2022)Stiller, Makdani, P{\"o}schel, Pausch, Debus,
  Bussmann, and Hoffmann]{stiller2022continual}
Stiller, P., Makdani, V., P{\"o}schel, F., Pausch, R., Debus, A., Bussmann, M.,
  and Hoffmann, N.
\newblock Continual learning autoencoder training for a particle-in-cell
  simulation via streaming.
\newblock \emph{ArXiv preprint}, abs/2211.04770, 2022.

\bibitem[Takamoto et~al.(2022)Takamoto, Praditia, Leiteritz, MacKinlay,
  Alesiani, Pfl{\"u}ger, and Niepert]{takamoto2022pdebench}
Takamoto, M., Praditia, T., Leiteritz, R., MacKinlay, D., Alesiani, F.,
  Pfl{\"u}ger, D., and Niepert, M.
\newblock Pdebench: An extensive benchmark for scientific machine learning.
\newblock In \emph{Thirty-sixth Conference on Neural Information Processing
  Systems Datasets and Benchmarks Track}, 2022.

\bibitem[Terraz et~al.(2017)Terraz, Ribes, Fournier, Iooss, and
  Raffin]{terraz2017melissa}
Terraz, T., Ribes, A., Fournier, Y., Iooss, B., and Raffin, B.
\newblock Melissa: large scale in transit sensitivity analysis avoiding
  intermediate files.
\newblock In \emph{Proceedings of the international conference for high
  performance computing, networking, storage and analysis}, pp.\  1--14, 2017.

\bibitem[Um et~al.(2020)Um, Brand, Fei, Holl, and Thuerey]{um2020solver}
Um, K., Brand, R., Fei, Y.~R., Holl, P., and Thuerey, N.
\newblock Solver-in-the-loop: Learning from differentiable physics to interact
  with iterative pde-solvers.
\newblock In Larochelle, H., Ranzato, M., Hadsell, R., Balcan, M., and Lin, H.
  (eds.), \emph{Advances in Neural Information Processing Systems 33: Annual
  Conference on Neural Information Processing Systems 2020, NeurIPS 2020,
  December 6-12, 2020, virtual}, 2020.

\bibitem[Wandel et~al.(2021)Wandel, Weinmann, and Klein]{wandel2020learning}
Wandel, N., Weinmann, M., and Klein, R.
\newblock Learning incompressible fluid dynamics from scratch - towards fast,
  differentiable fluid models that generalize.
\newblock In \emph{9th International Conference on Learning Representations,
  {ICLR} 2021, Virtual Event, Austria, May 3-7, 2021}, 2021.

\bibitem[Wang et~al.(2020)Wang, Kashinath, Mustafa, Albert, and
  Yu]{wang2020towards}
Wang, R., Kashinath, K., Mustafa, M., Albert, A., and Yu, R.
\newblock Towards physics-informed deep learning for turbulent flow prediction.
\newblock In Gupta, R., Liu, Y., Tang, J., and Prakash, B.~A. (eds.),
  \emph{{KDD} '20: The 26th {ACM} {SIGKDD} Conference on Knowledge Discovery
  and Data Mining, Virtual Event, CA, USA, August 23-27, 2020}, pp.\
  1457--1466. {ACM}, 2020.

\bibitem[Yoo et~al.(2003)Yoo, Jette, and Grondona]{yoo2003slurm}
Yoo, A.~B., Jette, M.~A., and Grondona, M.
\newblock Slurm: Simple linux utility for resource management.
\newblock In \emph{Workshop on job scheduling strategies for parallel
  processing}, pp.\  44--60. Springer, 2003.

\bibitem[You et~al.(2020)You, Li, Reddi, Hseu, Kumar, Bhojanapalli, Song,
  Demmel, Keutzer, and Hsieh]{You2020Large}
You, Y., Li, J., Reddi, S.~J., Hseu, J., Kumar, S., Bhojanapalli, S., Song, X.,
  Demmel, J., Keutzer, K., and Hsieh, C.
\newblock Large batch optimization for deep learning: Training {BERT} in 76
  minutes.
\newblock In \emph{8th International Conference on Learning Representations,
  {ICLR} 2020, Addis Ababa, Ethiopia, April 26-30, 2020}, 2020.

\end{thebibliography}
\bibliographystyle{icml2023}

\newpage
\appendix
\onecolumn

\section{Framework Details}
\label{sec:appendix:framework}
\subsection{Implementation}

The framework is mainly implemented in Python. The server supports data parallel training for Pytorch and Tensorflow models. Supporting an existing parallel solver (MPI+X parallelization supported) requires 1) instrumenting the code to call the API (for C, C++, Fortran or Python) enabling to connect and ship the data to the server, 2) instructing the server on how to perform the training and 3) defining the experimental design, i.e. how to draw the input parameter set for each instance.

This paper focuses on full online training, but the framework can start from a pre-trained model, and data provided for training can mix some read from files using a proxy client. This enables to reduce repetitive data generation during the hyper-parameterization process.

Communications are implemented with ZeroMQ \cite{hintjens2013zeromq}. ZeroMQ manages asynchronous data transfers, stored when necessary in internal buffers on the client and server sides to absorb network variability. If these buffers fill up the simulation is suspended. On the server side a thread associated to each GPU is dedicated to data reception and insertion into the memory buffer. Data formatting occurs at this stage when required. Another thread builds batches from the content of the buffer and trains the model with them.

\subsection{Fault Tolerance}

The framework is designed for application on large clusters. It is made resilient to different faults that can occur on such infrastructure. Failing clients are automatically restarted. The server keeps track of the received time steps $u_\lambda^t$ and discards already received ones. If the server fails, it is restarted with all the running clients. In case of launcher failure, the currently running clients will run up to completion and the server will finish and checkpoint. A manual restart of the full application is then needed to restart from that checkpoint. The scheduler of the cluster (e.g. Slurm \cite{yoo2003slurm}) comes with its own fault tolerance mechanism. A failing request of the scheduler is simply resubmitted later.

\subsection{Reproducibility}

The stochastic components of the framework (the model's weights initialization, the simulation parameter sampler, and the buffer policy) are seeded. The framework operates with configuration files which also contributes to the reproducibility of training experiments. However, the distributed execution on a cluster comes with inevitable variability which makes identical reproducibility challenging. Indeed,   the execution of the clients is subject to the workload of the cluster, which impacts the order of the data received by the server and may ultimately lead to variations in the training. 

\section{Equations}
\label{sec:appendix:equations}

\subsection{The Heat Equation}

\autoref{eq:heat} describes the evolution of the temperature $u$ in a 2D square
domain of length $L$. $\alpha$ represents the thermal diffusivity of the medium.
The system is fully determined by initial conditions (\autoref{eq:heat_ic}) and
boundary conditions (\autoref{eq:heat_bc}). The solver used to generate the
training data approximates the solution with an implicit Euler scheme of finite
difference. It solves the equation on a $100 \times 100$ regular grid, with time
steps of  $0.01$s for a total simulated time of $1$s. The thermal diffusivity is
fixed, $\alpha=1\ \text{m}^2\cdot\text{s}^{-1}$.  The initial temperature is
considered uniform over the domain, while the temperatures at the edge of the
domain are constant but not necessarily equal.  All temperature conditions take
random values in $[100\text{K}, 500\text{K}]$.

\begin{align}
    \label{eq:heat}
    &\frac{\partial u}{\partial t} = \alpha \Delta u \\ 
    &u(x, y, t=0) = T_{\text{IC}} \label{eq:heat_ic} \\
    &u(x=0, y, t) = T_{x_1}, \ u(x=L, y, t) = T_{x_2} \nonumber \\
    &u(x, y=0, t) = T_{y_1}, \  u(x, y=L, t) = T_{y_2} \label{eq:heat_bc}
\end{align}

\subsection{The Lorenz's System}
\label{sec:appendix:lorenz}

\autoref{eq:lorenz} describes the famous chaotic system introduced by Lorenz
that serves as a simplified climate model \cite{lorenz1963deterministic}. An
explicit Euler integration scheme allows the recovery of the trajectory of the
system given an initial position. It approximates the solution with time steps
of $0.01$s for a total simulated time of $20$s. For the dataset generation, the
parameters $\sigma$ and $\beta$ are fixed to the respective values of $10$ and 
$8/3$. The initial position of each trajectory is randomly taken according to
the normal distribution $\mathcal{N}(15, 30)$. $\rho$ takes values in $[0, 20,
40, 60, 80, 100]$.

\begin{align}
   \frac{dx}{dt} &= \sigma(y-z) \nonumber \\
   \frac{dy}{dt} &= x(\rho - z) - y \nonumber \\
   \frac{dz}{dt} &= xy - \beta z \nonumber \\
   \label{eq:lorenz}
\end{align}

\subsection{Simple Advection}

The Advection is a standard example of PDEs. The implementation of the example
is directly taken from the PDEBench paper \cite{takamoto2022pdebench} \S D.1.

\subsection{Mixed Advection-Diffusion}

The Advection-Diffusion considers a combination of phenomena: the simple advection, and the diffusion of a quantity $u$. The implementation of the example is directly taken from the original paper of the \textbf{Message Passing PDE Solver} \cite{brandstetter2021message} experiment \emph{E3} in \S 4.1.

\subsection{The Navier-Stokes Equations}

The Navier-Stokes equations are canonical to describe the evolution of any fluid
\cite{ferziger2002computational}. \autoref{eq:ns}, \ref{eq:ns_incompressible} 
and \ref{eq:ns_ic} describe the evolution of a 2D incompressible fluid
on the surface of the unit torus, in terms of velocity
$u$, and vorticity $w$ where  $w=\nabla \times u$. $\nu$ denotes the viscosity of the
fluid. It is set to $1\text{E}^{-5}\text{m}^2\cdot\text{s}^{-1}$. $f$ denotes
the forcing term of the system. The equations are solved with a pseudo-spectral
method for $(x, t) \in [0,1]^2 \times [0, T]$ and periodic boundary conditions.
The equations are discretized on a 64 x 64 regular mesh, with a time step of
$1\text{E}^{-4}s$ for a total simulated time of $20$s.  The solver generating
the training data is the same as for the original paper of \textbf{FNO}
\cite{li2020fourier}.

\begin{align}
    \label{eq:ns}
    & \partial_t w (x,t) + u(x,t)\cdot \nabla w(c,t) = \nu \Delta w(x,t) + f(x)  \\
    & \nabla \cdot u(x,t) = 0 \label{eq:ns_incompressible} \\
    & w(x, t=0) = w_0(x) \label{eq:ns_ic}
\end{align}

\section{Bias Mitigation}
\label{sec:appendix:bias_mitigation}

\begin{figure*}
    \centering
    \includegraphics[width=\textwidth]{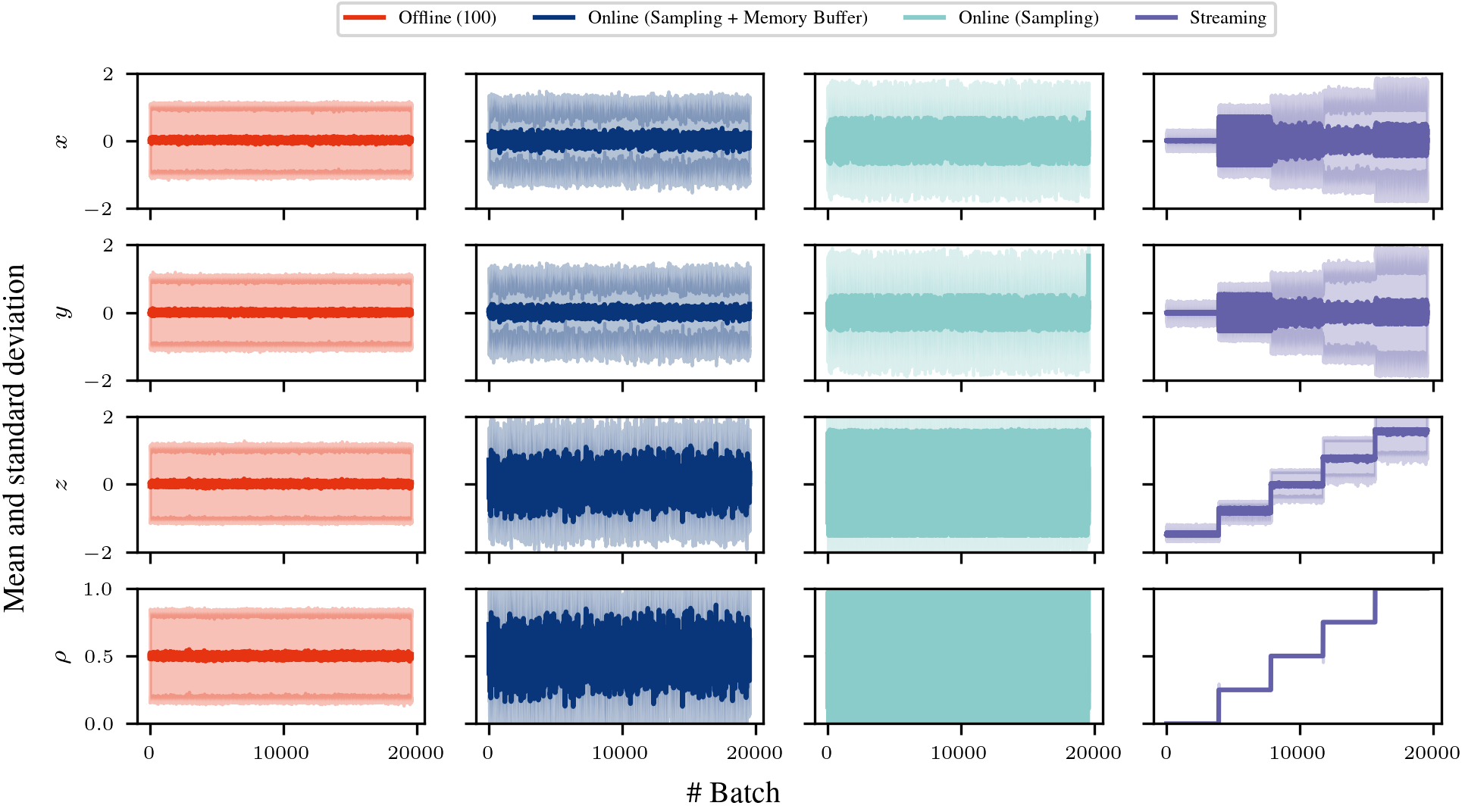}
    \caption{Statistics of the normalized inputs for the Lorenz's system (\hyperref[tab:experiments]{E2}) with different bias mitigation strategies.}
    \label{fig:batch_statistics}
\end{figure*}

The online setting of the framework inherently induces biases in the data available for training (\autoref{sec:datamanagement}). The efficiency of the bias mitigation strategy consisting of Monte Carlo sampling of the simulation parameters and mixing of the received data within the memory buffer is evaluated on the Lorenz's system described in \autoref{sec:appendix:lorenz}. \autoref{fig:batch_statistics} compares the batch statistics of the normalized inputs for different training settings identical to the ones presented in \autoref{sec:exp:sing_step}. $\rho$ is normalized between $0$ and $1$, using minimum and maximum values. Coordinates $x$, $y$, and $z$ are standardized. The Offline (100) corresponds to classical offline training with data sampled randomly from 100 previously generated trajectories. The sampling being uniform, the batches present an almost constant mean and standard deviation. It is assumed that representative and well-balanced batches with constant statistics will lead to faster stochastic gradient descent convergence and better generalization capabilities of the trained model. Streaming refers to the case where trajectories are generated by order of increasing $\rho$ and batches are formed with data as soon as they are received by the server. \autoref{fig:batch_statistics} highlights batch statistics are biased towards the simulation that has just been executed. For the online sampling, the parameter $\rho$ is randomly sampled. It immediately improves the fluctuations of batch statistics. The addition of the memory buffer further improves the batch statistics making them closer to the ones observed in the offline setting. 

\section{Prediction Results}

\begin{figure*}[!ht]
    \centering
    \subfigure{
        \includegraphics[width=0.5\textwidth]{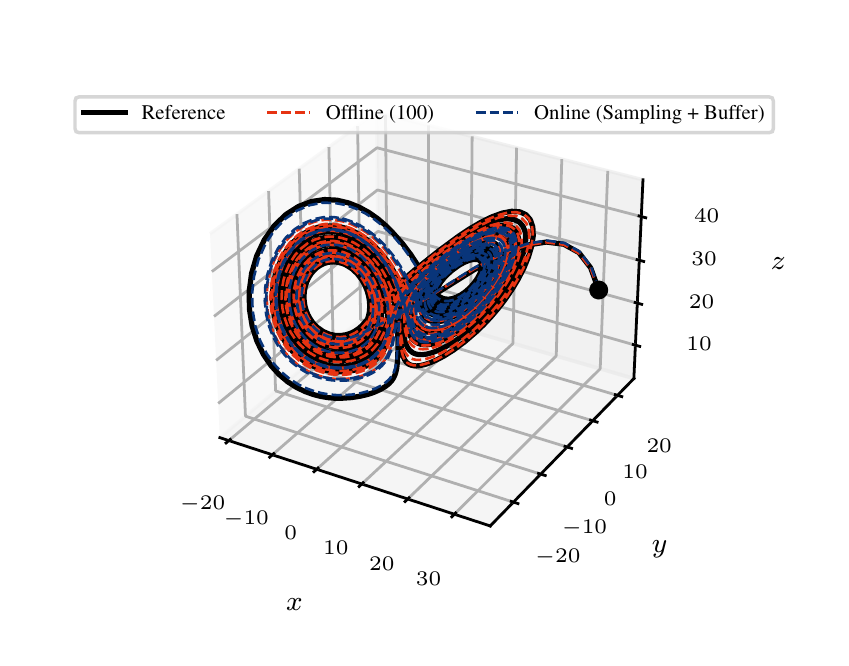}
    }%
    \subfigure{
        \includegraphics[width=0.5\textwidth]{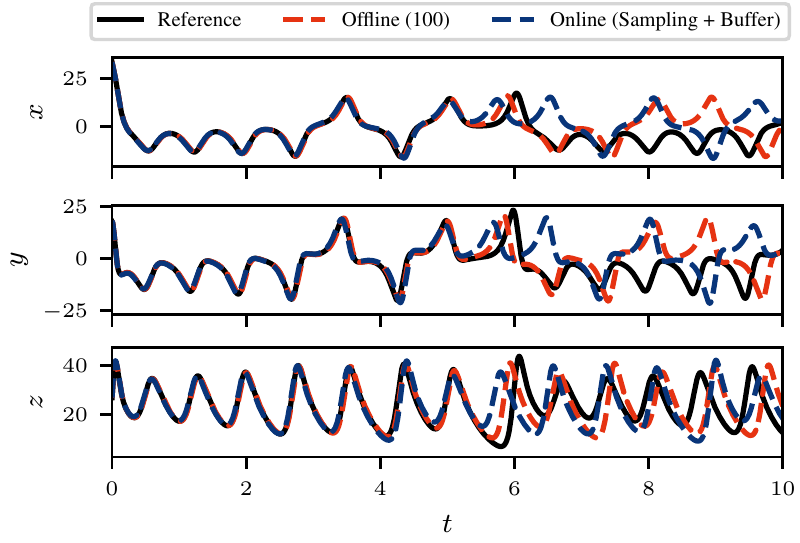}
    }
    \caption{Comparison of the predictions for the Lorenz's system and a trajectory never seen during training with $\rho=28$.}
    \label{fig:lorenz_predictions}
\end{figure*}

\begin{figure*}[!ht]
    \centering
    \includegraphics[width=\textwidth]{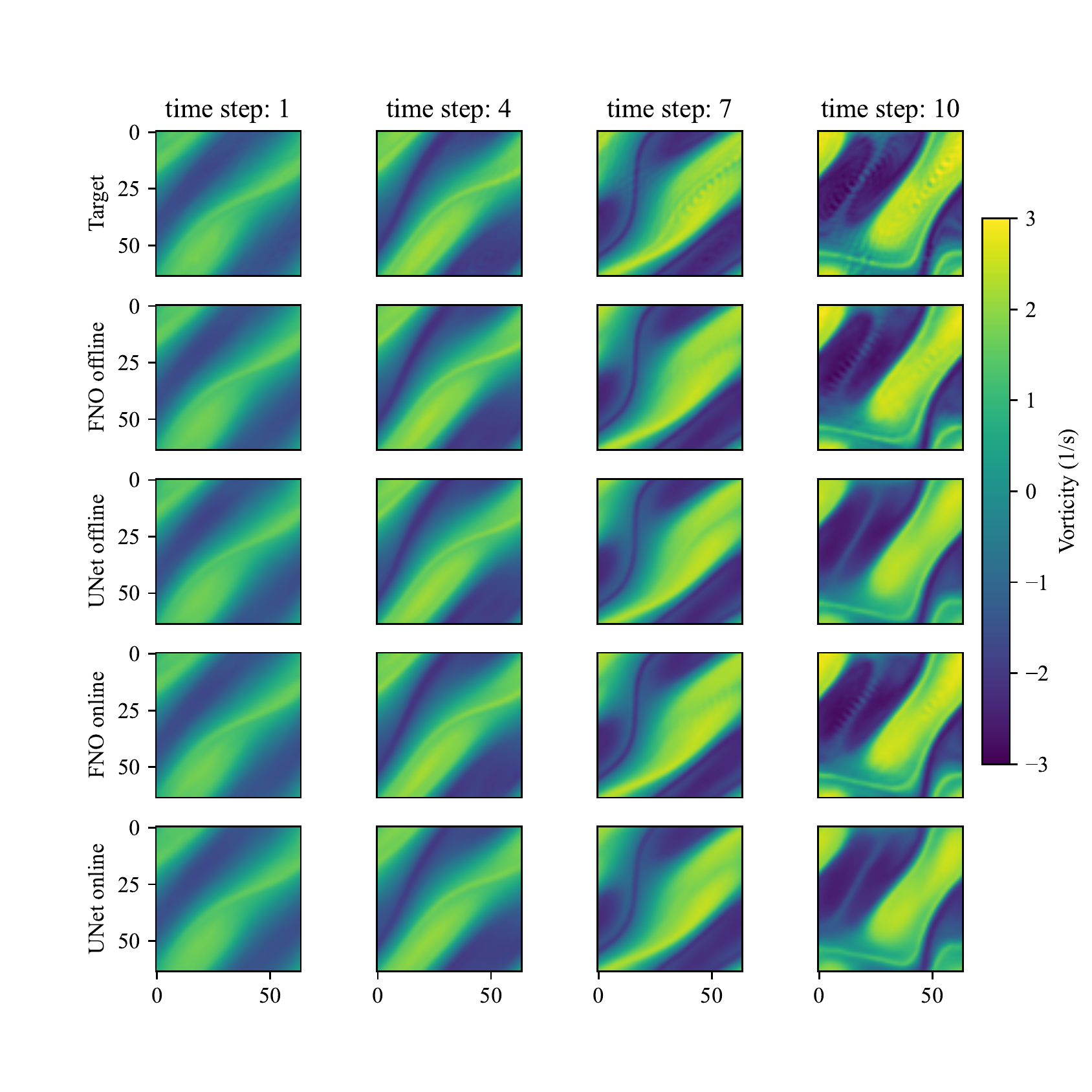}
    \caption{Visual comparison of Navier-Stokes predictions from U-Net and FNO}
    \label{fig:navierstokes_grid}
\end{figure*}

\begin{figure*}[!ht]
    \centering
    \includegraphics[width=\textwidth]{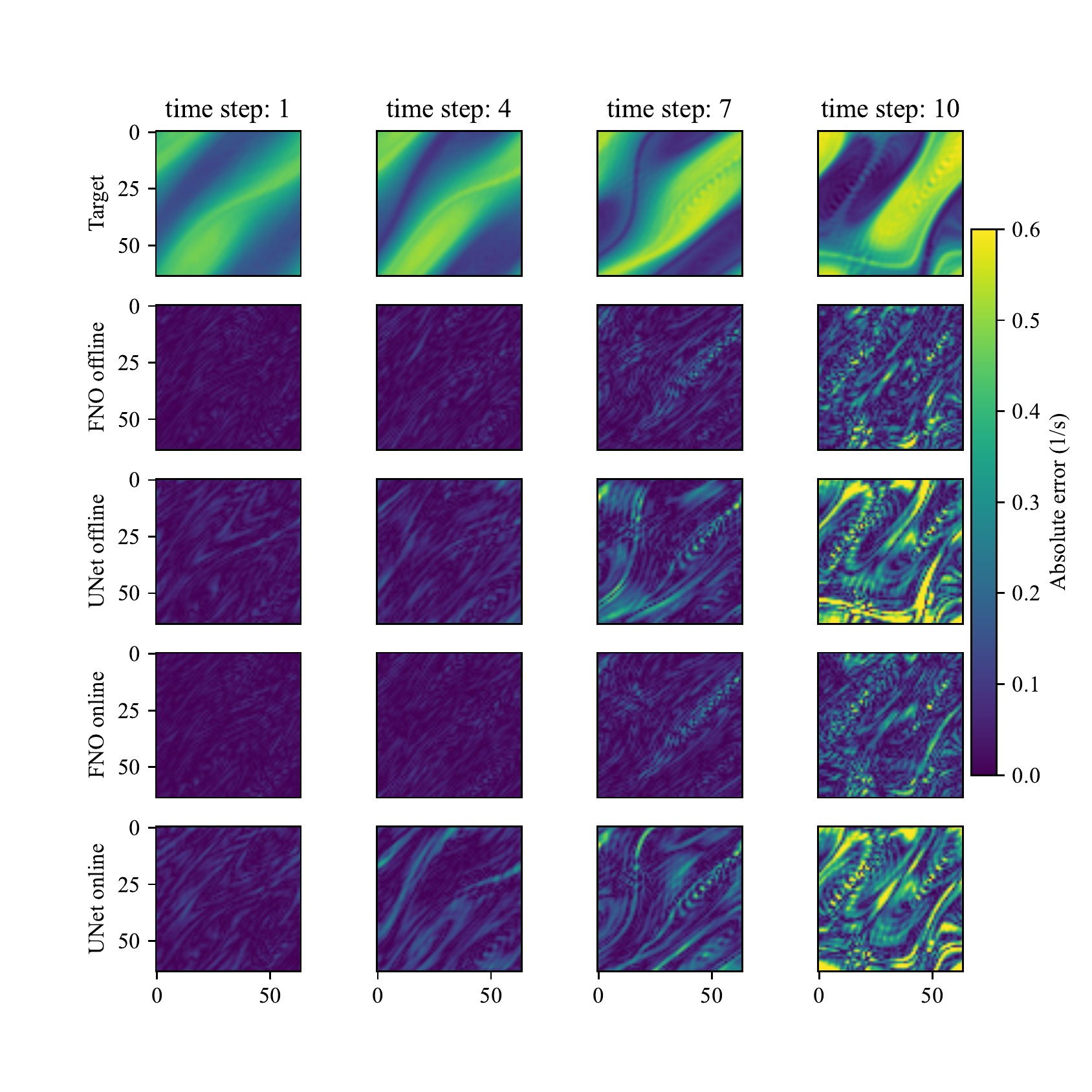}
    \caption{Visual comparison of Navier-Stokes prediction error from U-Net and FNO}
    \label{fig:navierstokes_grid_rr}
\end{figure*}

\begin{figure*}[ht]
    \centering
    \includegraphics[width=\textwidth]{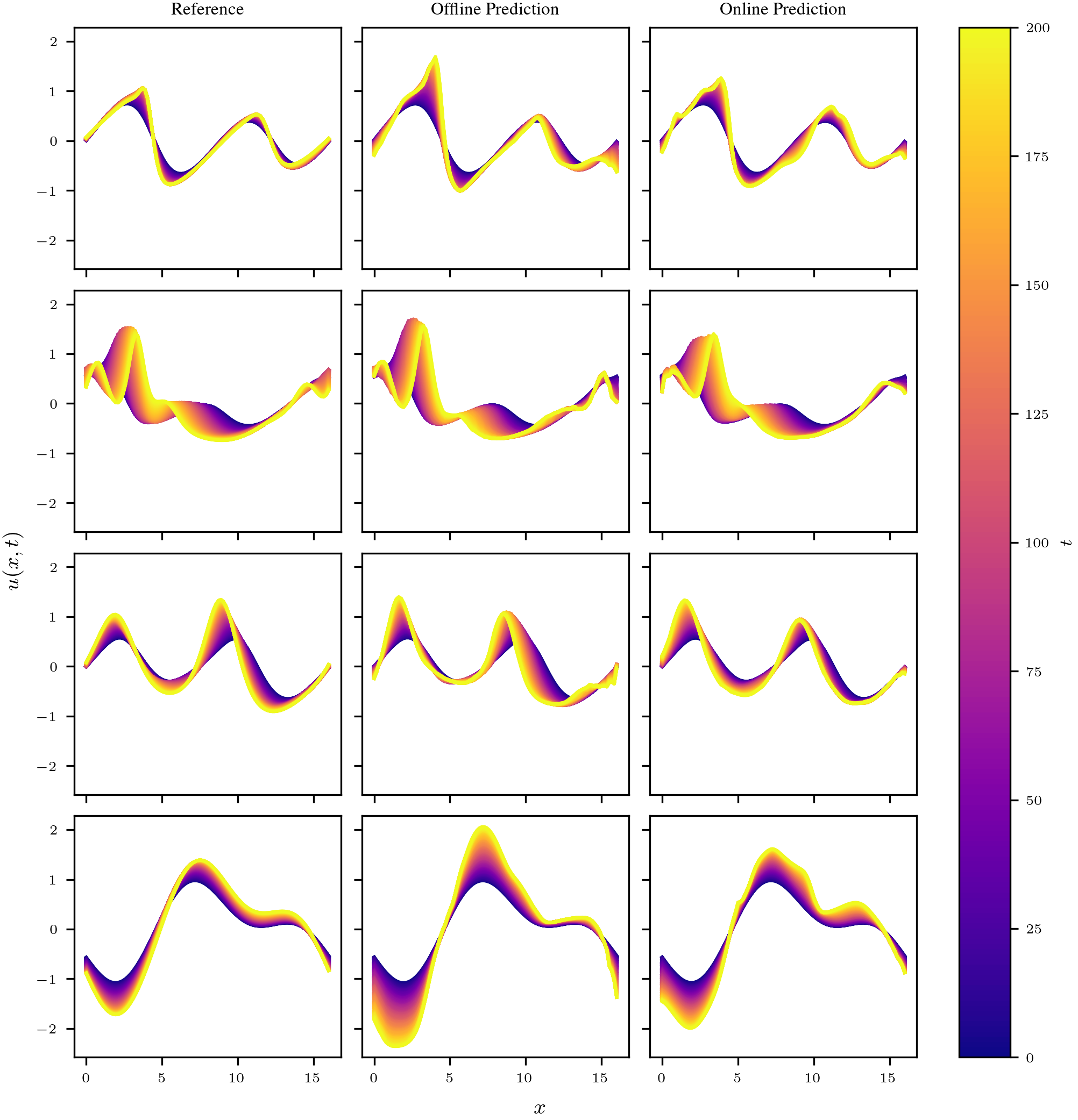}
    \caption{Visual comparison of Message Passing PDE Solver predictions on mixed advection-diffusion}
    \label{fig:mp_pde_grid}
\end{figure*}

\end{document}